\documentclass[conference]{IEEEtran}
\IEEEoverridecommandlockouts
\usepackage{cite}
\usepackage{amsmath,amssymb,amsfonts}
\usepackage{algorithmic}
\usepackage{graphicx}
\usepackage{textcomp}
\usepackage{xcolor}
\usepackage{mathrsfs}
\usepackage{amsthm}
\usepackage{diagbox}

\newtheorem{thm}{Theorem}[section]

\newtheorem{prop}{Proposition}[section]
\newtheorem{remark}{Remark}
\def\BibTeX{{\rm B\kern-.05em{\sc i\kern-.025em b}\kern-.08em
    T\kern-.1667em\lower.7ex\hbox{E}\kern-.125emX}}
\usepackage[top=1in,bottom=1in, right=0.8in,left=0.8in,nohead,nofoot]{geometry}
\begin{document}

\title{On Consistency of Graph-based Semi-supervised Learning \\
{\footnotesize}
}

%

\author{\IEEEauthorblockN{Chengan Du}
\IEEEauthorblockA{\textit{Center for Outcomes Research} \\ \textit{and Evaluation} \\
\textit{Yale University}\\
{New Haven, Connecticut}  \\
{chengan.du@yale.edu} }
\and
\IEEEauthorblockN{Yunpeng Zhao}
\IEEEauthorblockA{\textit{School of Mathematical and} \\ \textit{Natural Sciences} \\
\textit{ Arizona State University}\\
{Tempe, Arizona} \\
{ yunpeng.zhao@asu.edu} }
\and
\IEEEauthorblockN{Feng Wang}
\IEEEauthorblockA{\textit{ School of Mathematical and} \\ \textit{Natural Sciences} \\
{\textit{Arizona State University}}\\
{Tempe, Arizona} \\
{ fwang25@asu.edu}}

}

\maketitle
\begin{abstract}
Graph-based semi-supervised learning is one of the most popular methods in machine learning. Some of its theoretical properties such as bounds for the generalization error and the convergence of the graph Laplacian regularizer have been studied in computer science and statistics literature. However, a fundamental statistical property --\textit{consistency}\footnote{Throughout the paper, \textit{consistency} is used as a statistical term referring to an asymptotic property -- that is, the prediction by the algorithm  can identify the underlying truth with unlimited data. This is not to be confused with the existence of solutions in an equation system, which is a term used in algebra.} -- has not been proved. 

In this article, we study the consistency problem under a non-parametric framework. We obtain the following two results: 1) We prove that graph-based semi-supervised learning on the test data is consistent in the case that the estimated scores are enforced to be equal to the observed responses for the labeled data (the hard criterion). The sample size of unlabeled data are allowed to grow at a slower rate than the size of the labeled data in this result. 2) We give a counterexample demonstrating that the estimator can be inconsistent for the case when the estimated scores are not required to be equal to the observed responses (the soft criterion), where a tuning parameter is used to balance the loss function and the graph Laplacian regularizer. These somewhat surprising theoretical findings are supported by numerical studies on both synthetic and real datasets. 

Moreover, numerical studies show that the hard criterion constantly outperforms the soft criterion even when the sample size of unlabeled data is smaller than the size of labeled data. This suggests that practitioners can safely choose the hard criterion without the burden of selecting the tuning parameter in the soft criterion. 


\end{abstract}


\section{Introduction}

Semi-supervised learning is a class of machine learning methods in the middle ground between \textit{supervised learning}, where all training data are labeled, and \textit{unsupervised learning}, where no training data are labeled. Specifically, in addition to the labeled training data $X_1,\dots ,X_{n}$, there exist unlabeled inputs $X_{n+1},\dots ,X_{n+m}$. Under certain assumptions on the geometric structure of the input data, such as the cluster assumption or the low-dimensional manifold assumption \cite{ssl}, the use of both labeled and unlabeled data can achieve better prediction accuracy than supervised learning, which only uses labeled inputs $X_1,\dots ,X_{n}$.

Semi-supervised learning has become popular because the acquisition of unlabeled data is relatively inexpensive. A large number of methods have been developed under the framework of semi-supervised learning. For example, \cite{ratandven} proposed that the combination of labeled and unlabeled data will improve the prediction accuracy under the assumption of mixture models. The self-training method \cite{rose} and the co-training method \cite{jones} were than applied to semi-supervised learning when mixture models are not assumed. Reference \cite{c2}
described an approach to semi-supervised clustering based on hidden Markov random fields (HMRFs) that can combine multiple approaches in a unified probabilistic framework. Reference \cite{c3} proposed a probabilistic framework for semi-supervised learning incorporating a K-means-type clustering algorithm (HMRF-Kmeans). Reference \cite{c4} proposed the transductive support vector machines (TSVMs) that used the idea of transductive learning by including unlabeled data in the computation of the margin. 
Reference \cite{c6} used a convex relaxation of the optimization problem called semi-definite programming as a different approaches to the TSVMs.

In this article, we focus on a particular semi-supervised method -- graph-based semi-supervised learning. In this method, the geometric structure of the input data are represented by a weighted graph $\mathbf{G}=({V},{E})$, where nodes $V=\{v_1,\dots,v_{n+m}\}$ represent the inputs $X_1,\dots X_{n+m}$ and edges $E$ represent the similarities between them. The similarities are given in an $n+m$ by $n+m$ symmetric similarity matrix (or called \textit{kernel} matrix), $\mathbf{W}=[w_{ij} ]$, where $0 \leq w_{ij} \leq 1$. The larger $w_{ij}$ implies that $X_i$ and $X_j$ are more similar. Furthermore, let $Y_1,\dots,Y_n$ be the responses of the labeled data.

Reference \cite{zhu2003} proposed the following graph-based learning method,
\begin{align}\label{hard}
\min_{\mathbf{f}=(f_1,\dots,f_{n+m})^T} \sum_{i=1}^{n+m} \sum_{j=1}^{n+m} w_{ij}(f_i-f_j)^2
\end{align}
\begin{center}
	subject to  $f_i=Y_i, i=1, \dots,n$.
\end{center}
We call the solution estimated scores. The objective function \eqref{hard} (hereafter referred to as ``hard criterion''), requires all estimated scores to be exactly the same as the responses for the labeled data. Reference \cite{d2005} relaxed this requirement by proposing a soft version (hereafter referred to as ``soft criterion''). We follow an equivalent form given in \cite{zhuandb},
\begin{align}\label{soft}
& \min_{\mathbf{f}=(f_1,\dots,f_{n+m})^T} \sum_{i=1}^n (Y_i-f_i)^2 \nonumber \\
& \quad \quad \quad  + \frac{\lambda}{2} \sum_{i=1}^{n+m} \sum_{j=1}^{n+m} w_{ij}(f_i-f_j)^2.
\end{align}
The soft criterion belongs to the ``loss+penalty'' paradigm: it searches for the minimizer $\mathbf{\hat{f}}$, which improves the smoothness of $\mathbf{\hat{f}}$ by a penalty-based similarity matrix while causing a training error. 

These two criteria are closely related: when $\lambda=0$ the soft criterion is equivalent to the hard criterion.

\begin{remark}
	The tuning parameter $\lambda$ being 0 in the soft criterion \eqref{soft} is understood in the following sense: the squared loss has an infinite weight and thereby $f_i$ is enforced to be $Y_i$ for all the labeled data. But the penalty term $\sum_{i=1}^{n+m} \sum_{j=1}^{n+m} w_{ij}(f_i-f_j)^2$ still plays a crucial role when it has no conflict with the hard constraints on the labeled data -- that is, it builds a connection between $f_i$'s on the labeled and unlabeled data. More precisely, the solution of Eq. \eqref{soft} goes to the solution of Eq. \eqref{hard} as $\lambda \rightarrow 0$ (see page 203 of \cite{ssl} and Proposition \ref{propo_equivalent} for a detailed explanation).

\end{remark}

References \cite{zhou2004, b2004}  also proposed different variants of graph-based learning methods. We only focus on Eq. \eqref{hard} and \eqref{soft} in this article.

The theoretical properties of graph-based learning have been studied in computer science and statistics literature. Reference \cite{007} derived the limit of the Laplacian regularizer when the sample size of unlabeled data goes to infinity. Reference \cite{008} considered the convergence of Laplacian regularizer on Riemannian manifolds. Reference \cite{005} reinterpreted the graph Laplacian as a measure of intrinsic distances between inputs on a manifold and reformulated the problem as a functional optimization in a reproducing kernel Hilbert space. Reference \cite{toyota} pointed out that the hard criterion can yield a completely noninformative solution when the size of unlabeled data goes to infinity and labeled data are finite -- that is, the solution can give a perfect fit on the labeled data but remains as 0 on the unlabeled data. Reference \cite{wass} obtained the asymptotic mean squared error of a different version of graph-based learning criterion. Reference \cite{b2004} gave a bound of the generalization error for a slightly different version of objective function \eqref{soft}. Reference \cite{pmlr-v49-elalaoui16} studied the theoretical properties of $\ell_p$-based Laplacian regularization -- in particular the phase transition of $p$ for an informative solution.

However, no result is available in the literature on a very fundamental question -- the consistency of graph-based learning -- which is the main focus of this article. Specifically, we want to answer the question of \textit{under what conditions $\hat{f}_i$ will converge to $\mathbb{E}[Y_i|X_i]$ on unlabeled data,} where $\mathbb{E}[Y_i|X_i]$ is the true probability of a positive label given $X_i$ if responses are binary, and $\mathbb{E}[Y_i|X_i]$ is the regression function on $X_i$ if responses are continuous. For simplicity, we will always call $\mathbb{E}[Y_i|X_i]$ as regression function.

Most of the literatures on graph-based semi-supervised learning considered a ``functional version'' of Eq. \eqref{hard} and \eqref{soft}. Previous works used a functional optimization problem with the optimizer $\hat{f}(x)$ being a function, as an approximation of the original problem with the optimizer $\mathbf{\hat{f}}$ being a vector. The behavior of the limit of graph Laplacian and the solution $\hat{f}(x)$ were studied in this context. 

Instead of adopting this framework, we use a more direct approach. We focus on the original problem and study the relations of $\hat{f}_i$ and $\mathbb{E}[Y_i|X_i]$ directly under the general non-parametric setting. Our approach essentially belongs to the framework of transductive learning, which focuses on the prediction on the given unlabeled data $X_{n+1},\dots, X_{n+m}$, not the general mapping from inputs to responses. By establishing a link between the optimizer of Eq. \eqref{hard} and the Nadaraya-Watson estimator \cite{nada,watson} for kernel regression, we prove the consistency of the hard criterion. Unlike \cite{toyota}, our result requires the sample size of labeled data goes infinity, which is natural in asymptotic theory of statistics. The result also allows the size of unlabeled data goes to infinity. On the other hand, we show that the soft criterion is inconsistent for sufficiently large $\lambda$. To the best of our knowledge, this is the first result that explicitly distinguishes the hard criterion and the soft criterion of graph-based learning from a theoretical perspective. 

The main results are stated in Section \ref{sec:main} and proved in Section \ref{sec:proof}. We give a toy example in Section \ref{sec:toy} to give more intuition into the somewhat surprising theoretical findings. The results are further supported by numerical studies on synthetic and real datasets in Section \ref{sec:sim}. Moreover, numerical studies also show that the hard criterion constantly outperforms the soft criterion even when the sample size of unlabeled data is smaller than the size of labeled data. This suggests that practitioners can safely choose the hard criterion without the burden of selecting the tuning parameter in the soft criterion. 

\section{MAIN RESULTS}\label{sec:main}
Let $(X_1,Y_1), \dots ,(X_{n+m},Y_{n+m})$ be independently and identically distributed pairs. Each $X_i$ is a $d$-dimensional vector and $\mathbf{Y}=(Y_1,\dots,Y_{n+m})^T$ are binary responses labeled as 1 and 0 (the classification case) or continuous responses (the regression case). The last $m$ responses are unobserved.

We now give the solution of the soft version \eqref{soft}. 
Recall that $W$ is the similarity matrix. Let
$ \bf{D} $ $=\textnormal{diag}(d_1,\dots,d_{n+m})$ where $d_i=\sum \limits_{j=1}^{n+m} w_{ij}$, and $\mathbf{L}=\mathbf{D}-\mathbf{W}$ being the unnormalized graph Laplacian (see \cite{Newman2010} for details). Soft criterion \eqref{soft} can be written in matrix form
\begin{align}\label{matrix}
\min_{\mathbf{f}} (\mathbf{f}-\mathbf{Y})^T {\bf V}(\mathbf{f}-\mathbf{Y})+ \lambda \mathbf{f}^T \mathbf{L} \mathbf{f},
\end{align}
where $\bf V$ is an $n+m$ by $n+m$ matrix defined as
\begin{align*}
\bf{V} &=\begin{pmatrix}
\mathbf{I}_n & \bf{0} \nonumber \\
\bf{0} & \bf{0}
\end{pmatrix}. \nonumber
\end{align*}
By taking the derivative of Eq. \eqref{matrix} with respect to $\mathbf{f}$ and setting equal to zero, we obtain the following solution:
\begin{align*}\nonumber
\mathbf{\hat{f}} & =(\mathbf{V}+\lambda \mathbf{L})^{-1}\begin{pmatrix}
\mathbf{Y}_n  \nonumber\\
\mathbf{0}
\end{pmatrix} \nonumber.
\end{align*}
where $\mathbf{Y}_n=(Y_1,\dots,Y_n)^T$.

Our objective is to determine the estimated scores on the unlabeled data, i.e., $\mathbf{\hat{f}}_{(n+1):(n+m)}=(\hat{f}_{n+1},\dots,\hat{f}_{n+m})^T$. In order to obtain an explicit form for $\mathbf{\hat{f}}_{(n+1):(n+m)}$, we use a formula for inverse of a block matrix: for any non-singular square matrix
\[
\bf A=\begin{pmatrix}
\mathbf{A}_{11} & \mathbf{A}_{12} \nonumber \\
\mathbf{A}_{21} & \mathbf{A}_{22}
\end{pmatrix}, \]
\begin{align*}
\mathbf{A}^{-1} = &  \left (  \begin{tabular}{c}
$(\mathbf{A}_{11}-\mathbf{A}_{12}\mathbf{A}_{22}^{-1}\mathbf{A}_{21})^{-1}$  \\
$-(\mathbf{A}_{22}-\mathbf{A}_{21}\mathbf{A}_{11}^{-1}\mathbf{A}_{12})^{-1}\mathbf{A}_{21}\mathbf{A}_{11}^{-1}$
\end{tabular}
\right . 
\end{align*}
\vspace{-0.15in}
\begin{align*}
\quad \quad \quad \quad & \left . \begin{tabular}{c}
$-(\mathbf{A}_{11}-\mathbf{A}_{12}\mathbf{A}_{22}^{-1}\mathbf{A}_{21})^{-1}\mathbf{A}_{12}\mathbf{A}_{22}^{-1} $  \\
$(\mathbf{A}_{22}-\mathbf{A}_{21}\mathbf{A}_{11}^{-1}\mathbf{A}_{12})^{-1} $
\end{tabular}
\right ).
\end{align*}
Write $\bf D$ and $\bf W$ as $2 \times 2$ block matrices,
\begin{align*} \nonumber
\bf D=\begin{pmatrix}
\mathbf{D}_{11} & \mathbf{D}_{12} \nonumber \\
\mathbf{D}_{21} & \mathbf{D}_{22}
\end{pmatrix},
\bf W=\begin{pmatrix}
\mathbf{W}_{11} & \mathbf{W}_{12} \nonumber \\
\mathbf{W}_{21} & \mathbf{W}_{22}
\end{pmatrix}.
\end{align*}
By the formula above,
\begin{align} 
& \mathbf{\hat{f}}_{(n+1):(n+m)}= \nonumber \\
& ( \mathbf{D}_{22}-\mathbf{W}_{22}-\lambda \mathbf{W}_{21} (\mathbf{I}_n+\lambda \mathbf{D}_{11}-\lambda \mathbf{W}_{11})^{-1}\mathbf{W}_{12})^{-1} \nonumber \\
&  \quad \cdot \mathbf{W}_{21}(\mathbf{I}_n+\lambda \mathbf{D}_{11}-\lambda \mathbf{W}_{11})^{-1} {\mathbf{Y}_n}. \label{solution_soft}
\end{align}
By letting $\lambda=0$, we obtain
\begin{align}\label{solution_hard}
\mathbf{\hat{f}}_{(n+1):(n+m)}=( \mathbf{D}_{22}-\mathbf{W}_{22})^{-1}  \mathbf{W}_{21} {\mathbf{Y}_n}.
\end{align}
Some literatures such as \cite{zhuandb} used Lagrange multipliers to the constrained optimization problem and obtained the same solution for the hard criterion \eqref{hard}. Therefore, we have essentially proved the following proposition:
\begin{prop}\label{propo_equivalent}
	The solution of the soft criterion \eqref{soft} at $\lambda=0$ is equivalent to the solution of the hard criterion.
\end{prop}

It is worth noting that the time complexities of computing Eq. \eqref{solution_hard} and \eqref{solution_soft} are respectively $O(m^3)$ and $O((m+n)^3)$ when using Gaussian elimination for solving systems of linear equations. Therefore, it is more efficient to solve the hard criterion, which is another advantage of the hard criterion in addition to theoretical considerations.

The form of Eq. \eqref{solution_hard} is closely related to the Nadaraya-Watson estimator \cite{nada,watson} for kernel regression, which is
\begin{align}\label{NW}
\hat{q}_{n+a}= \frac{\sum_{i=1}^n w_{n+a,i}Y_i}{\sum_{k=1}^n w_{n+a,k}}, \quad a=1, \dots, m.
\end{align}

The Nadaraya-Watson estimator is well studied under the non-parametric framework. We can construct $\bf W$ by a kernel function -- that is, let $w_{ij}=K((X_i-X_j)/h_n)$, where $K$ is a nonnegative function on $\mathbb{R}^d$, and $h_n$ is a positive constant controlling the bandwidth of the kernel. Let $q(X)=\mathbb{E}[Y|X]$ be the true regression function. The consistency of Nadaraya-Watson estimator was first proved by \cite{watson,nada}. Many other researchers such as \cite{nwcon} and \cite{nwcon1} studied its asymptotic properties under different assumptions. Here, we follow the result in \cite{dandw}. If $h_n \to 0$, $nh_n^d \to \infty$ as $n \to \infty$, and $K$ satisfies:
\begin{itemize}
	\vspace{0.1in}
	\item [(i)] $K$ is bounded by $k^* < \infty$;
	\item [(ii)] The support of $K$ is compact;
	\item [(iii)] $K \ge \beta I_{B}$ for some $\beta >0$ and some closed ball $B$ centered at the origin and having positive radius $\delta$,
	\vspace{0.1in}
\end{itemize}
then $\hat{q}_{n+a}$ converges to $q(X_{n+a})$ in probability for $a=1,\dots, m$.

By establishing a connection between the solution of the hard criterion and Nadaraya-Watson estimator, we prove the following main theorem:
\begin{thm} \label{th1}
	Suppose that \\ $(X_1,Y_1),\dots ,(X_{n+m},Y_{n+m})$ are independently and identically distributed with $Y_i$ being bounded; $h_n$ and $K$ satisfy the above conditions. Further, we assume that the density function $\phi(\cdot)$ of $X_1$ has a compact support $\mathscr{X}$. And for every inner point $x$ in $\mathscr{X}$,
	\begin{align} \label{try}
	\phi(x) \ge s^*>0.
	\end{align}	
	Then, for $m =o(nh_n^d)$, $\hat{f}_{n+a}$ given in Eq. \eqref{solution_soft} converges to $q(X_{n+a})$ in probability, for $a=1,\dots, m$.
\end{thm}

The proof will be given in Section \ref{sec:proof}.

Theorem \ref{th1} establishes the consistency of the hard criterion under the standard non-parametric framework with two additional assumptions. First, both labeled data and unlabeled data are allowed to grow but the size of unlabeled data $m$ grows slower than the size of labeled data $n$. We conjecture that when $m$ grows faster than $n$, the graph-based semi-supervised learning may not be consistent based on the simulation studies in Section \ref{sec:sim} although it still outperforms the soft criterion. Reference \cite{toyota} also suggested that the method may not work well when $m$ grows too fast. 
Second, we assume the density function of the difference of
two independent inputs is strictly positive near the origin, which is a mild technical condition valid
for commonly used density functions.

We now consider the soft criterion ($\lambda \neq 0$).
\begin{prop}\label{soft_result}
	Suppose that \\
	$(X_1,Y_1),\dots ,(X_{n+m},Y_{n+m})$ are independently and identically distributed with $Y_i$ being bounded. Furthermore, suppose $\bf W$ represents a connected graph. Then for sufficiently large $\lambda$, the soft criterion \eqref{soft} is inconsistent.
\end{prop}
\begin{proof}
	Consider another extreme case of the soft criterion \eqref{soft}, $\lambda=\infty$. When $\bf W$ represents a connected graph, the objective function becomes
	\begin{align}\label{1987}
	\min_{\mathbf{f}=(f_1,\dots,f_n)^T} \sum_{i=1}^{n} (Y_i-f_i)^2
	\end{align}
	\begin{center}
		subject to  $f_i=f_j, 1 \le i,j \le n+m$.
	\end{center}
	It is simple to verify that the solution of Eq. \eqref{1987}, denoted by $\mathbf{\hat{f}}(\infty)$, is given by
	\begin{align*}
	\hat{f}_{n+a}(\infty)=\frac{1}{n} \sum_{i=1}^n Y_i, \,\, a=1,\dots, m.
	\end{align*}
	By the law of large numbers,
	\begin{align*}
	\lim_{n \to \infty} \hat{f}_{n+a}(\infty)=\mathbb{E}[q(X_{1})] \,\, \mbox{almost surely}.
	\end{align*}
	Clearly, $\mathbb{E}[q(X_{1})] \neq q(X_{n+a})$ since the right-hand side is a random variable. This implies that for sufficiently large $\lambda$, the soft criterion is inconsistent.
\end{proof}

This counterexample in fact suggests more general results than its first look. We prove that under $\lambda \rightarrow \infty$, the soft criterion predicts the same label (an extremely inaccurate prediction). On the contrary, the hard criterion $(\lambda=0)$ gives a consistent prediction. Note that Eq. \eqref{solution_hard} is a continuous function of $\lambda$, so the prediction cannot suddenly jump from consistent to extremely inaccurate. This suggests in general the soft criterion is inconsistent for all or a wide range of nonzero $\lambda$.

\section{A TOY EXAMPLE}{\label{sec:toy}}

Theorem \ref{th1} and Proposition \ref{soft_result} provide somewhat surprising insights about the graph-based semi-supervised learning. At a first glance, the hard criterion makes an impractical assumption that requires the responses to be noiseless, while the soft criterion seems to be a more natural choice. According to our theoretical result, the hard criterion is however consistent under the standard non-parametric framework where the responses on training data are allowed to be random and noisy by default. Below we provide a toy example\footnote{This example is motivated by comments from an anonymous reviewer.} that further illustrates the rationale behind the hard criterion. 

Consider the case of $X_1,\cdots, X_{n+m}$ being the same constant. Thus, $Y_1,\cdots,Y_{n+m}$ become independently and identically distributed random variables. We still assume that the first $n$ responses are observed and the last $m$ are to be predicted. Let $\bf W$ be the Gaussian radial basis function (RBF) kernel, that is,
\begin{align}\nonumber
w_{ij}=\exp \left (-\frac{\|X_i-X_j\|^2}{\sigma^2} \right ), \,\, \mbox{for } 1 \le i,j \le m+n.
\end{align}
Then $w_{ij} \equiv 1$ for $1 \le i,j \le m+n$. 
Thus, 
\begin{align*}
& \mathbf{D}_{22}-\mathbf{W}_{22}= \\
& \begin{pmatrix}
m+n-1 & -1 & \cdots & -1	\\
-1 & m+n-1 & \cdots & -1 \\
\vdots & \vdots & \ddots & \vdots \\
-1 & -1 & \cdots & m+n-1
\end{pmatrix}_{m\times m}.
\end{align*}
It is easy to verify that\footnote{Note that the size of $\mathbf{D}_{22}-\mathbf{W}_{22}$ is $m \times m$ but not $(m+n)\times (m+n)$ and this matrix is invertible. } 
\begin{align*}
& (\mathbf{D}_{22}-\mathbf{W}_{22})^{-1} \\
& = \begin{pmatrix}
\frac{\strut n+1}{ \strut n(m+n)} & \frac{\strut 1}{\strut n(m+n)} & \cdots &  \frac{\strut 1}{\strut n(m+n)}	\\
\frac{\strut 1}{\strut n(m+n)} & 	\frac{\strut n+1}{\strut n(m+n)} & \cdots & \frac{\strut 1}{\strut n(m+n)} \\
\vdots & \vdots & \ddots & \vdots \\
\frac{\strut 1}{\strut n(m+n)} & \frac{\strut 1}{\strut n(m+n)} & \cdots & \frac{\strut n+1}{\strut n(m+n)}
\end{pmatrix}_{m \times m}.
\end{align*}
Thus, from Eq. \eqref{solution_hard} the solution of the hard criterion is $\mathbf{\hat{f}}_{i} \equiv \frac{1}{n} \sum_{j=1}^n Y_j$ for $i=n+1,...,n+m$, and $ \mathbf{\hat{f}}_{i} =Y_i$ for $i=1,...,n$. This is in fact the best solution one can expect: for labeled data we simply used the observed responses and for unlabeled data we used the mean of the observed responses to do prediction (since there is no other information available). This example shows in the transductive learning setup it is not an issue that the hard criterion did not smoothen the labeled data (there is no need to do prediction on labeled data in the first place), while the prediction on unlabeled data is indeed based on a weighted average of the observed responses.

\section{PROOF OF THE MAIN THEOREM}\label{sec:proof}

We give the proof of Theorem \ref{th1} in this section.

Recall that
\begin{align}\nonumber
\mathbf{\hat{f}}_{(n+1):(n+m)}={(\mathbf{D}_{22}-\mathbf{W}_{22})^{-1}\mathbf{W}_{21}}{\mathbf{Y}_n}.
\end{align}
We first focus on $(\mathbf{D}_{22}-\mathbf{W}_{22})^{-1}$. Clearly,
\begin{align*}
(\mathbf{D}_{22}-\mathbf{W}_{22})^{-1}=(\mathbf{I}_m-\mathbf{D}_{22}^{-1}\mathbf{W}_{22})^{-1}\mathbf{D}_{22}^{-1} \nonumber .
\end{align*}
For any positive integer $l$, define
\begin{align}
{\bf S}_l = & {\mathbf{D}_{22}^{-1}\mathbf{W}_{22}}+({\mathbf{D}_{22}^{-1}\mathbf{W}_{22}})^2+({\mathbf{D}_{22}^{-1}\mathbf{W}_{22}})^3+\dots \nonumber \\
& +({\mathbf{D}_{22}^{-1}\mathbf{W}_{22}})^l \nonumber.
\end{align}
Our goal is to prove that the limit of $S_l$
exists with probability approaching 1, and thus we have
\begin{align}
{(\mathbf{I}_m-\mathbf{D}_{22}^{-1}\mathbf{W}_{22})^{-1}}={\mathbf{I}_m} + \lim_{l\to \infty} {\bf S}_l \nonumber
\end{align}
with probability approaching 1 \cite{proof}. \\

By definition, \\

$\mathbf{D}_{22} =
\left( {\begin{array}{*{20}c}
	d_{n+1,n+1} & \cdots & 0 \\ \vdots & \ddots & \vdots \\ 0 & \cdots & d_{n+m,n+m}
	\end{array} } \right),$
\vspace{0.1in}

$\mathbf{W}_{22} =
\left( {\begin{array}{*{20}c}
	w_{n+1,n+1} & \cdots & w_{n+1,n+m} \\ \vdots & \ddots & \vdots \\ w_{n+m,n+1} & \cdots & w_{n+m,n+m}
	\end{array} } \right),$\\
where
\begin{align}\nonumber
d_{n+a,n+a}=\sum\limits _{k=1}^{n+m} w_{n+a,k}, \quad w_{n+a,i}=K \left ( \frac{X_{i}-X_{n+a}}{h_n} \right ),
\end{align}
for $1 \le a \le m, 1 \le i \le n+m$.
Thus we have
\begin{align*}
&\mathbf{D}_{22}^{-1}\mathbf{W}_{22} = \\
&	\begin{pmatrix}
	\frac{\strut w_{n+1,n+1}}{\strut d_{n+1,n+1}} & \cdots & \frac{\strut w_{n+1,n+m}}{\strut d_{n+1,n+1}} \\ \vdots & \ddots & \vdots \\ \frac{\strut w_{n+1,n+m}}{\strut d_{n+m,n+m}} & \cdots & \frac{\strut w_{n+m,n+m}}{\strut d_{n+m,n+m}}
	\end{pmatrix}.
\end{align*}

Define $p(X_{n+a}) = \mathbb{P}(\|X_i-X_{n+a}\|\le \delta h_n \mid X_{n+a})$. Since $h_n \to 0$, there exist $n_0 \in \mathbb{N}$ such that $\delta h_n \le c$ holds for every $n>n_0$. Then by Eq. \eqref{try} and the definition of multiple integral, with probability 1.
\begin{align*}
\lim_{n\rightarrow \infty} \frac{p(X_{n+a})}{V_d(\delta h_n)} = \phi(X_{n+a}) \geq s^* ,
\end{align*}
where $V_d(\delta h_n)$ denotes the volume of a $d$-dimensional ball with radius $\delta h_n$. Thus, for sufficiently large $n$,
\begin{align*}
p(X_{n+a}) \ge \frac{1}{2}s^*V_d(\delta h_n)= s h_n^d,
\end{align*}
where $s$ is a constant only related to $s^*$ and $\delta$.

Since $n h_n^d \to \infty$, the above inequality implies $n p(X_{n+a})  \to \infty$. On the other side, $p(X_{n+a})  \to 0$ since $h_n \to 0$.

Further,
\begin{align*}\nonumber
&\text{Var}(I\{\|X_i-X_{n+a}\|\le \delta h_n\} \mid X_{n+a}) \\
&=p(X_{n+a})(1-p(X_{n+a}) ).
\end{align*}
By Chebyshev's Inequality, for any $0<\epsilon<1/2$, since $nh_n^d \to \infty$,
\begin{align}
& \mathbb{P} \left ( \left . \left |\frac{\sum _{i=1}^nI\{\|X_i-X_{n+a}\|\le \delta h_n\} }{np(X_{n+a}) }-1 \right |\ge \epsilon  \right | X_{n+a}  \right) \nonumber \\[5pt]
=& \mathbb{P} \left (\left . \left |\frac{1}{n}\sum \limits_{i=1}^nI\{\|X_i-X_{n+a}\|\le \delta h_n\}-p(X_{n+a})  \right |\ge \right.  \right. \nonumber \\
& \quad \left. \epsilon p(X_{n+a})  \mid X_{n+a} \right ) \nonumber \\[5pt]
\le &\frac{p(X_{n+a}) (1-p(X_{n+a}) )}{n\epsilon^2 p(X_{n+a}) ^2} \le \frac{1}{\epsilon^2 p(X_{n+a}) n} \le \frac{1}{\epsilon^2snh_n^d}. \label{try1}
\end{align}
Therefore,
\begin{align*}
& \mathbb{P} \left (\left |\frac{\sum _{i=1}^nI\{\|X_i-X_{n+a}\|\le \delta h_n\} }{np(X_{n+a}) }-1 \right |\ge \epsilon   \right) \nonumber \\
& \le \frac{1}{\epsilon^2snh_n^d} \to 0 \quad \textnormal{as } n \to \infty.
\end{align*}
This further implies
\begin{align*}\nonumber
\frac{\sum \limits_{i=1}^nI\{\|X_i-X_{n+a}\|\le \delta h_n\} }{np(X_{n+a})  } \to 1 \quad \textnormal{in probability.}
\end{align*}
We now continue to study the property of $\mathbf{D}_{22}^{-1}\mathbf{W}_{22}$. Consider each element ${(\mathbf{D}_{22}^{-1}\mathbf{W}_{22})}_{ab}$ of this matrix. For $1 \le a, b \le m$,
\begin{align*}\nonumber
& {(\mathbf{D}_{22}^{-1}\mathbf{W}_{22})}_{ab} = \frac{w_{n+a,n+b}}{d_{n+a,n+a}} \nonumber \\[5pt]
& =K \left (\frac{X_{n+b}-X_{n+a}}{h_n} \right)/\sum \limits _{i=1}^{n+m}K \left(\frac{X_i-X_{n+a}}{h_n} \right )\\[5pt]
&\le \frac{k^*}{\beta\sum\limits_{i=1}^n I\{\|X_i-X_{n+a}\|\le \delta h_n\}}, \nonumber
\end{align*}
by condition (i) and (iii). For simplicity of notation, let $$ \Phi_n(a)=\frac{\sum \limits_{i=1}^nI\{\|X_i-X_{n+a}\|\le \delta h_n\} }{np(X_{n+a}) },$$ where $\Phi_n$ is a nonnegative function depending on $n$.
By Eq. \eqref{try1}, we have
\begin{align}
\mathbb{P} (0 \le \Phi_n(a)\le 1-\epsilon) \le
\mathbb{P} (|\Phi_n(a)-1|\ge \epsilon)\le\frac{1}{\epsilon^2snh_n^d}, \nonumber
\end{align}
which implies
\begin{align}\nonumber
& \mathbb{P} \left (\min_{1\le a \le m}\Phi_n(a)\le 1-\epsilon \right ) = \mathbb{P} \left (\bigcup\limits_{a=1}^{m}\{\Phi_n(a)\le 1-\epsilon\} \right ) \\
&\le \sum \limits_{a=1}^{m} \mathbb{P}(\Phi_n(a)\le 1-\epsilon) \le \frac{m}{\epsilon^2snh_n^d}  ,\nonumber
\end{align}
and
\begin{align*}
& \mathbb{P} \left (\max\limits_{1\le a \le m}\frac{k^*}{\beta \Phi_n(a)np(X_{n+a}) }
\le \frac{k^*}{\beta(1-\epsilon)np(X_{n+a})} \right ) \\[5pt]
& \ge 1-\frac{m}{\epsilon^2snh_n^d} .
\end{align*}
Since $\frac{m}{\epsilon^2snh_n^d}\to 0$, we have
\begin{align}\label{try_2}
&\mathbb{P} \left (\max\limits_{1\le a,b \le m}( {{\mathbf{D}_{22}^{-1}\mathbf{W}_{22})}_{ab}}\le \max\limits_{1\le a \le m}\frac{k^*}{\beta \Phi_n(a)np(X_{n+a}) } \right. \nonumber  \\
& \quad \left. \le  M\frac{1}{nh_n^d} \right ) \to 1  , \quad \textnormal{as } n \to \infty,
\end{align}
where $M=\frac{2k^*}{s\beta}>\frac{k^*}{(1-\epsilon)s\beta}$. Note that $M$ is a constant independent with $n$ and $m$.

For the sake of simplicity, we say a matrix $\bf A$ has \textit{tiny elements}, if
\begin{align} \nonumber
{\bf\|A\|_{\text{max}}} \le M\frac{1}{nh_n^d},
\end{align}
with probability approaching 1, where ${\bf\| A\|_{\max}}=\max_{ij} {\bf A}_{ij}$. ${\bf (A)}_i$ denotes the $i$-th row of $\bf A$. Thus,  $\mathbf{D}_{22}^{-1}\mathbf{W}_{22}$ has tiny elements by Eq. \eqref{try_2}. Moreover,
\begin{align}\nonumber
\|({\mathbf{D}_{22}^{-1}\mathbf{W}_{22}})^2\|_{\text{max}}&= \|(\mathbf{D}_{22}^{-1}\mathbf{W}_{22})(\mathbf{D}_{22}^{-1}\mathbf{W}_{22})\|_{\text{max}}  \nonumber \\
&\le(M \frac{1}{nh_n^d})^2 m= \frac{M}{nh_n^d} (\frac{mM}{nh_n^d} ) \nonumber
\end{align}
holds with probability approaching 1. By induction,
\begin{align*} \nonumber
& { \|({\mathbf{D}_{22}^{-1}\mathbf{W}_{22}})^l\|_{\text{max}}}= { \|({\mathbf{D}_{22}^{-1}\mathbf{W}_{22}})({\mathbf{D}_{22}^{-1}\mathbf{W}_{22}})^{l-1}\|_{\text{max}}} \\
& \le \frac{M}{nh_n^d} (\frac{mM}{nh_n^d} )^{l-1},
\end{align*}
\textnormal{with probability approaching 1}.
Therefore,
\begin{align}
\|{\bf S}_l\|_{\text{max}} =&  \|{\mathbf{D}_{22}^{-1}\mathbf{W}_{22}}+\dots+({\mathbf{D}_{22}^{-1}\mathbf{W}_{22}})^l\|_{\text{max}} \nonumber \\
\leq & \|{\mathbf{D}_{22}^{-1}\mathbf{W}_{22}}\|_{\text{max}}+\dots+\|({\mathbf{D}_{22}^{-1}\mathbf{W}_{22}})^l\|_{\text{max}} \nonumber \\
\le& \frac{M}{nh_n^d} \left (1+\dots+(\frac{mM}{nh_n^d})^{l-1} \right )    \nonumber
\end{align}
with probability approaching 1.
\begin{align}
\lim_{l\to \infty} \|{\bf S}_l\|_{\text{max}} &\le \lim_{l\to \infty} \frac{M}{nh_n^d} \left (1+\dots+(\frac{mM}{nh_n^d})^{l-1} \right) \nonumber \\[5pt]
&\leq \frac{M}{nh_n^d}/(1-\frac{mM}{nh_n^d}) \le \frac{2M}{nh_n^d}   \nonumber
\end{align}
with probability approaching 1.

Thus, ${\bf S }\stackrel{\triangle}{=} \underset{l \to \infty}{\lim} {\bf S}_l$ exists with probability approaching 1 since $\underset{l \to \infty}{\lim} \|{\bf S}_l\|_{\text{max}} < \infty$, and $\bf S$ also has tiny elements. Therefore,
\begin{align*}
(\mathbf{D}_{22}-\mathbf{W}_{22})^{-1}= & (\mathbf{I}_m-\mathbf{D}_{22}^{-1}\mathbf{W}_{22})^{-1}\mathbf{D}_{22}^{-1}\\
= & (\mathbf{I}_m + \mathbf{S})\mathbf{D}_{22}^{-1}, \nonumber
\end{align*}
with probability approaching 1.

We now go back to the solution of the hard criterion of graph-based semi-supervised learning,
\begin{align*}\nonumber
{\mathbf{\hat{f}}_{(n+1):(n+m)}} & =(\mathbf{D}_{22}-\mathbf{W}_{22})^{-1}\mathbf{W}_{21}\mathbf{Y}_n \\
&=(\mathbf{I}_m + \mathbf{S})\mathbf{D}_{22}^{-1}\mathbf{W}_{21}{\mathbf{Y}_n} \\
& =\mathbf{D}_{22}^{-1}\mathbf{W}_{21}{\mathbf{Y}_n}+\mathbf{SD}_{22}^{-1}\mathbf{W}_{21}{\mathbf{Y}_n},
\end{align*}
with probability approaching 1. For $1 \le a \le m$, $\hat{f}_{(n+a)}$ equals to the $a$th row of $(\mathbf{D}_{22}-\mathbf{W}_{22})^{-1}\mathbf{W}_{21}\mathbf{Y}_n$, i.e.,
\begin{align*}
\hat{f}_{(n+a)}&=\left( (\mathbf{D}_{22}-\mathbf{W}_{22})^{-1}\mathbf{W}_{21}\mathbf{Y}_n\right)_a \\[5pt]
&=\sum \limits _{i=1}^n \frac{w_{i,n+a}}{d_{n+a,n+a}} Y_i+ {\bf (S)}_{a}\mathbf{D}_{22}^{-1}\mathbf{W}_{21}\mathbf{Y}_n,\nonumber
\end{align*}
with probability approaching 1, where ${\bf (S)}_a$ denotes the $a$th row of $\bf S$.

By assumption, $Y_i$s are bounded. Without loss of generality, assume $\|Y_n\|_{\text{max}}\le 1$. For $1 \le a \le m$, define
\begin{align}
g_{(n+a)}=&\sum \limits _{i=1}^n Y_i \left (\frac{w_{i,n+a}}{\sum_{k=1}^n w_{k,n+a}}-\frac{w_{i,n+a}}{d_{n+a,n+a}} \right ). \nonumber
\end{align}
We have
\begin{align}
| g_{(n+a)}|\le& \sum \limits _{i=1}^n  \|Y_n\|_{\text{max}} \left (\frac{w_{i,n+a}}{\sum_{k=1}^n w_{k,n+a}}-\frac{w_{i,n+a}}{d_{n+a,n+a}} \right )  \nonumber \\[5pt]
=&\frac{\sum_{i=1}^n w_{i,n+a}}{\sum_{k=1}^n w_{k,n+a}}-\frac{\sum_{i=1}^n w_{i,n+a}}{\sum_{k=1}^{n+m} w_{k,n+a}} \nonumber\\[5pt]
=& \frac{\sum_{k=n+1}^{n+m} w_{k,n+a}}{ d_{n+a,n+a}}\nonumber\\[5pt]
\le &  \frac{mk^*}{\beta\Phi_n(a)np(X_{n+a}) } \le \frac{mM}{nh_n^d} \to 0, \nonumber
\end{align}
with probability approaching 1 as $n \to \infty$. This implies
\begin{align}\nonumber
g_{(n+a)} \to 0 \,\, \textnormal{in probability},
\end{align}
since for any $\epsilon>0$ we can find $m, n \in \mathbb{N}$ such that $\frac{mM}{nh_n^d} \le \epsilon$ and
\begin{align}\nonumber
\mathbb{P}(|g_{(n+a)}| \le \epsilon)\geq \mathbb{P} \left (|g_{(n+a)}| \le \frac{mM}{nh_n^d} \right )\to 1.
\end{align}

Finally, for each $1 \le a \le m$,
\begin{align*}
\hat{f}_{(n+a)}=&\sum \limits _{i=1}^n \frac{w_{i,n+a}}{d_{n+a,n+a}} Y_i+{\bf (S)}_{a}\mathbf{D}_{22}^{-1}\mathbf{W}_{21}\mathbf{Y}_n \nonumber \\[5pt]
=&\sum \limits _{i=1}^n \frac{w_{i,n+a}}{ \sum_{k=1}^{n} w_{k,n+a} } Y_i + {\bf (S)}_{a}\mathbf{D}_{22}^{-1}\mathbf{W}_{21}\mathbf{Y}_n \\
& - g_{(n+a)}\nonumber,
\end{align*}
Since $\bf S$ has tiny elements,
\begin{align}\nonumber
\|{\bf (S)}_{a}\mathbf{D}_{22}^{-1}\mathbf{W}_{21}\mathbf{Y}_n \| \leq \frac{mM}{nh_n^d}\to 0 \,\, 
\end{align}
with probability approaching 1,
which implies
$ {\bf (S)}_{a}\mathbf{D}_{22}^{-1}\mathbf{W}_{21}\mathbf{Y}_n\to 0$ \textnormal{in probability}.
The theorem then holds by the consistency of Nadaraya-Watson estimator.

\section{EXPERIMENTS}\label{sec:sim}
\subsection{Synthetic Data}
In this sub-section, we compare the performance of the hard and soft criteria with different tuning parameters under a linear and non-linear model.

The inputs $X_1,\dots X_{n+m}$ are generated independently from a truncated multivariate normal distribution. Specifically, let $\tilde{X}_i$ follow a $p$-dimensional multivariate normal with the mean $\mu=(0.5,\dots, 0.5)$ and the variance-covariance matrix
\[
\begin{pmatrix}
0.1 & 0.05 & 0.05 & \dots  & 0.05 \\
0.05 & 0.1 & 0.05 & \dots  & 0.05 \\
\vdots & \vdots & \vdots & \ddots & \vdots \\
0.05 & 0.05 & 0.05 & \dots  & 0.1
\end{pmatrix}.
\]
We set $p=5$. For $i=1,\dots, n+m$ and $k=1,\dots, p$, let $X_{ik}=\tilde{X}_{ik}$ if $\tilde{X}_{ik} \in [0,1]$ and $\tilde{X}_{ik}=0$ otherwise, where $X_{ik}$ and $\tilde{X}_{ik}$ are the $k$-th component of $X_i$ and $\tilde{X}_i$, respectively.

Let $\bf W$ be the Gaussian radial basis function (RBF) kernel with $\sigma=h_n=(\log n/n)^{1/5}$. Note that $\bf W$ has compact support since $X_i$ are truncated and the choice of $h_n$ satisfies the condition in Theorem \ref{th1}.

We consider two models in this study. In Model 1, the responses $Y_i$ follow a logistic regression with
\begin{align}\nonumber
\text{logit }q(X_i)= & -1.35+2X_{i1}-X_{i2}+X_{i3}-X_{i4} \\
& +2X_{i5},
\end{align}
for $i=1,\dots,m+n$. Model 2 uses a non-linear logit function,
\begin{align*}
\text{logit }q(X_i)= & -1.35+2X_{i1}-X_{i2}+X_{i3}-X_{i4} \\
& +2X_{i5}+X_{i1}X_{i3}+X_{i2}X_{i4},
\end{align*}
for $i=1,\dots,m+n$.

We compare the performance of graph-based learning methods with four different tuning parameters, $\lambda=0,0.01,0.1$ and 5. Performance is measured by the root mean squared error (RMSE) on the unlabeled data:
$$\sqrt{\frac{1}{m} \sum_{a=1}^m (q(X_{n+a})-\hat{q}_{n+a})^2}.$$ Each simulation is repeated 1000 times and the average RMSEs are reported.

Figure 1 shows the RMSEs under Model 1 when the sample size of unlabeled data $m$ is fixed as 30 and the sample size of labeled data $n=10$, 30, 50, 100, 200, 300, 500, 800, 1000 and 1500. As $n$ increases, the RMSEs of all methods decrease as expected. More importantly, the RMSE increases as $\lambda$ increases. In particular, the hard criterion always outperforms the soft criterion, which is in line with our theoretical results.

Figure 2 shows the RMSEs under Model 1 when $n$ is fixed as 100 and $m=30$, 60, 100, 300, 500 and 1000.  As before, the RMSE  increases as $\lambda$ increases. Moreover, the RMSEs of all methods increase as $m$ increases, which suggests that the hard criterion may not be consistent when $m$ grows faster than $n$ although the hard criterion still performs constantly better. For a non-linear logit function, Figure 3 and 4 show the same patterns as in Figure 1 and 2, which further supports our theoretical results.
\begin{figure}[h!]
	\centering
	\includegraphics[width=3.3in]{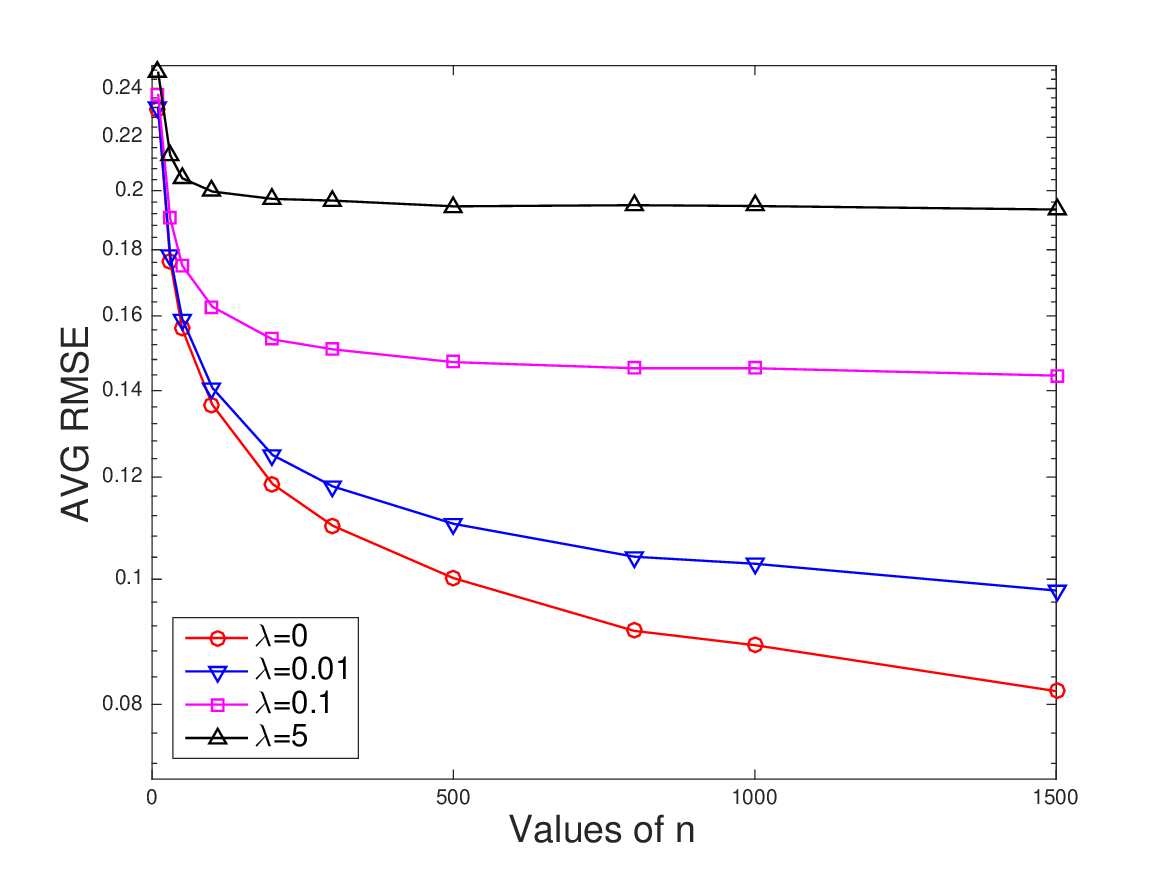}
	\caption{Average RMSEs for $m=30$ under Model 1}
\end{figure}
\begin{figure}[h!]
	\centering
	\includegraphics[width=3.3in]{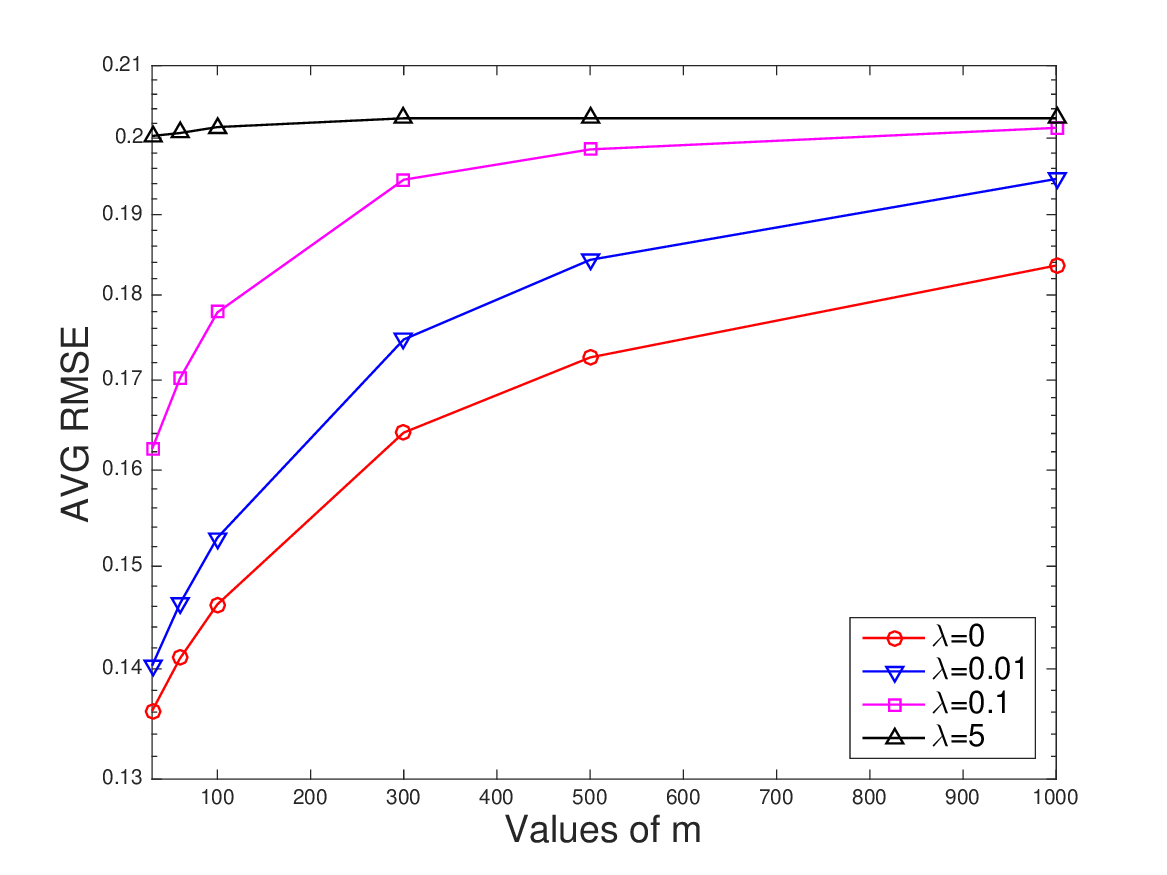}
	\caption{Average RMSEs for $n=100$ under Model 1}
\end{figure}
\begin{figure}[h!]
	\centering
	\includegraphics[width=3.3in]{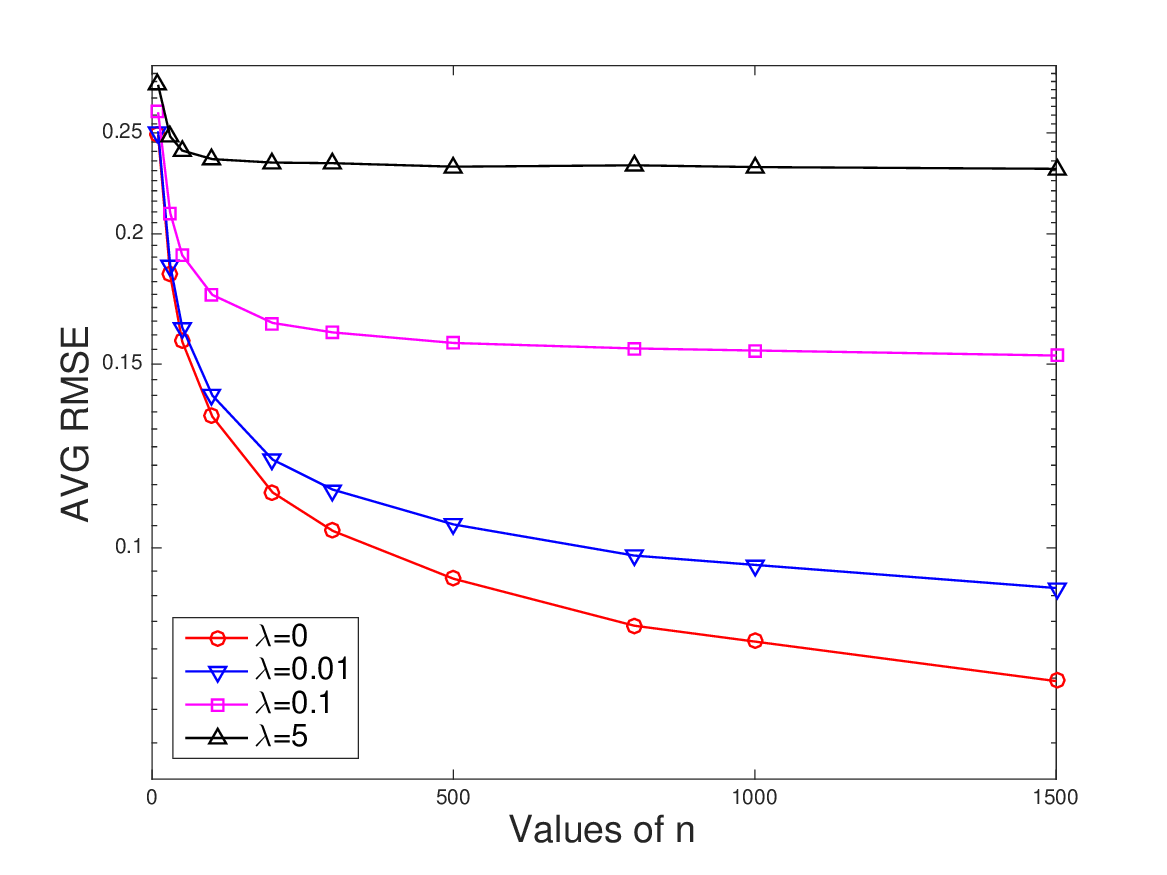}
	\caption{Average RMSEs for $m=30$ under Model 2}
\end{figure}
\begin{figure}[h!]
	\centering
	\includegraphics[width=3.3in]{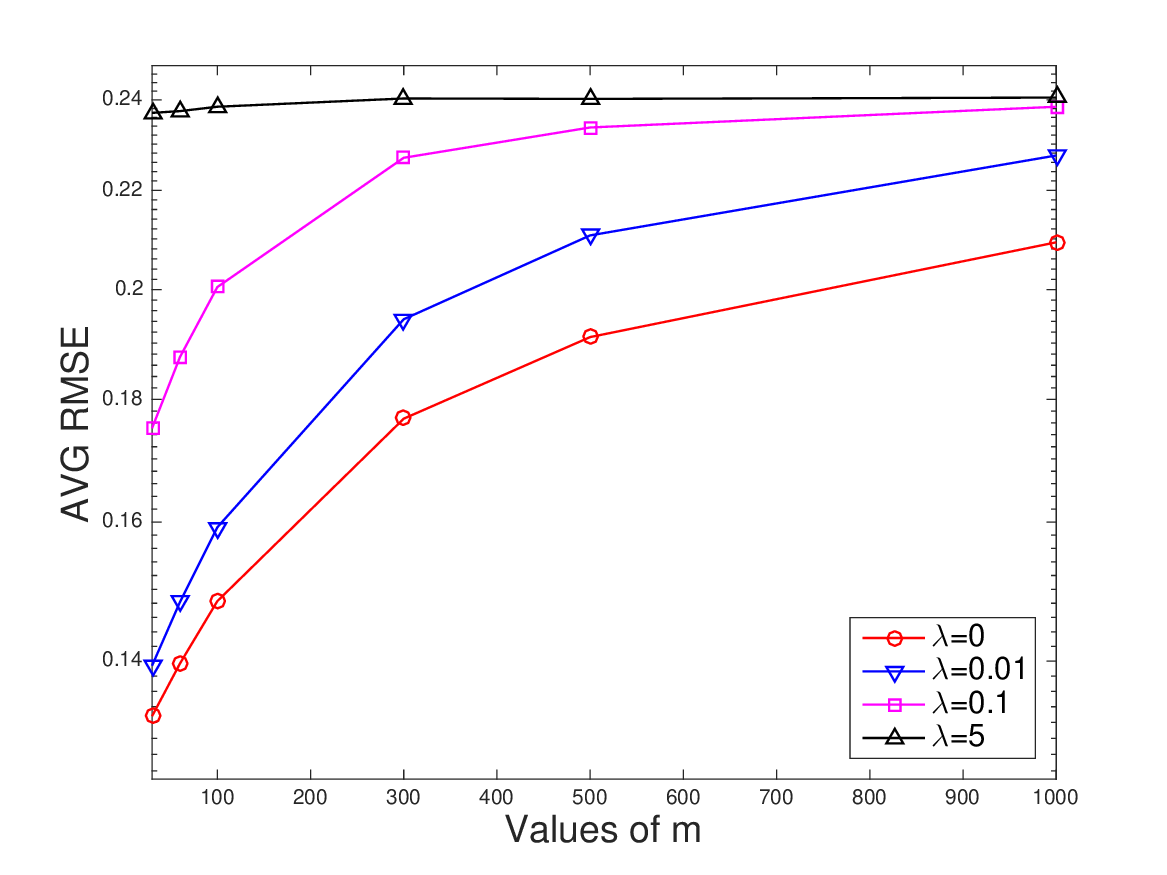}
	\caption{Average RMSEs for $n=100$ under Model 2}
\end{figure}

\subsection{The Columbia Object Image Library Data}
We test the performance of the hard and soft criteria on the Columbia object image library dataset compiled by \cite{ssl}, which is listed in the Chapter 21 of the book as a benchmark. The dataset contains color images of 24 different objects taken from 72 different angles. These subjects were classified into six classes and the authors randomly discarded 38 images of each class, leaving 250 each, i.e., 1500 samples in total. The author also created a binary version of this data, which groups the first three and last three together, respectively, leaving two classes. We use this binary dataset to test the hard and soft criteria. The inputs $X$ were created from $16\times 16$ pixels of each image. We use the Gaussian RBF kernel as $\mathbf{W}$ with $\sigma^2$ being the median of squared distances between each pair of inputs. 

When responses are binary, the RMSE cannot be used to measure the performance of classification algorithms in real dataset because the true probability $\mathbb{E}[Y_i|X_i]$ is unknown (for continuous responses, the root mean square prediction error can be used). Instead, we use the area under the receiver operating characteristic curve (AUC) to measure the performance. The receiver operating characteristic (ROC) curve is obtained via plotting the sensitivity (true positive rate) versus $1-$specificity ({false positive rate}) of the classification.

We vary the ratio between the training/labeled and test/unlabeled sets and prepare the data sets in the following way: in the first setting, we randomly split the data into five subsets of approximately equal size. We then use each subset as the test set and the rest four as the training set. In this way, every part of the data has the chance to be predicted in the experiment. We repeat the above procedure 100 times and thus the reported results are based on the average of 500 experiments. In each experiment, we compare the criteria on seven different tuning parameters, $\lambda=0,0.01,0.05,0.1,0.5,1$ and 5. 

In the second and the third settings, we follow the same procedure as above but use only 20\% and 10\% of the data as labeled samples, respectively. Specifically, we randomly split  the data into five subsets of approximately equal size and use one subset as the  training set and the rest four as the test set in the second setting. In the third setting, we split  the data into ten subsets and use one as the training and nine as the test. We repeat the above procedure 100 times in each setting. The reported results are thus based on the average of 500 experiments as before in the second setting but are the average of 1000 experiments in the last setting since we split the data into ten subsets under this scenario.

\begin{figure}[h!]
	\centering
	\includegraphics[width=3.3in]{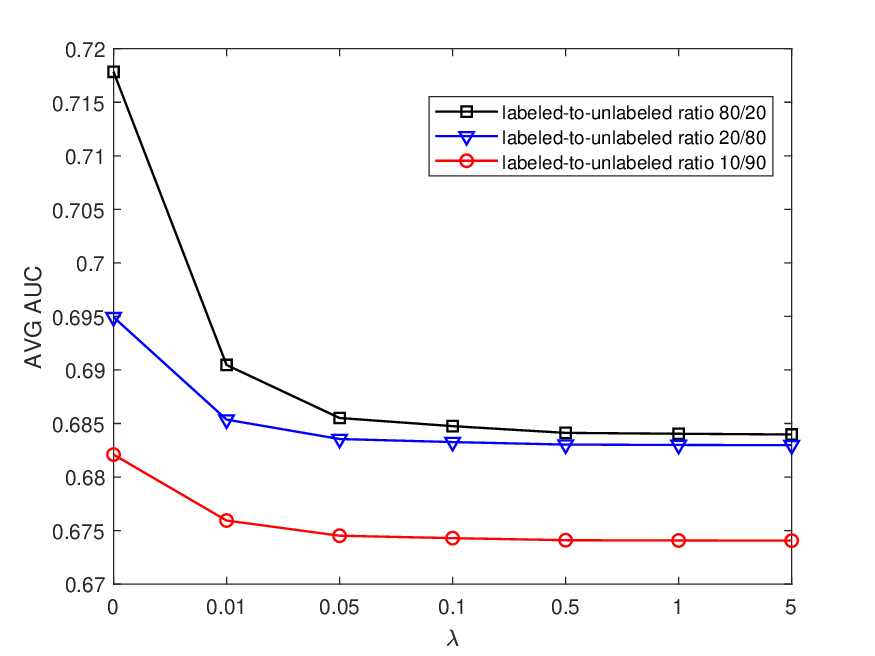}
	\caption{Average AUCs for the COIL data}
\end{figure}


According to Figure 5 the hard criterion ($\lambda=0$) constantly gives the best AUC on all combinations of the scales of the labeled and unlabeled data. The AUC decreases as $\lambda$ increases in general, although the difference between the AUCs for $\lambda=1$ and $\lambda=5$ are negligible. Moreover, the AUC decrease as the proportion of labeled data decreases as expected. The pattern is consistent with that of RMSEs for  the synthetic data and the theoretical results. 


\section{SUMMARY}\label{sec:summary}

In this article, we proved the consistency of graph-based semi-supervised learning when the tuning parameter of the graph Laplacian is zero (the hard criterion) and showed that the method can be inconsistent when the tuning parameter is nonzero (the soft criterion). Moreover, the numerical studies also suggest that the hard criterion outperforms the soft criterion in terms of RMSE and AUC. These results provide a better understanding about the statistical properties of graph-based semi-supervised learning. It suggests that practitioners can safely choose the hard criterion ($\lambda=0$) with no need for tuning $\lambda$ in the soft criterion. 

For future work, we plan to investigate the theoretical properties of other indicators of prediction accuracy such as AUC and MCC (Matthews correlation coefficient) in more depth. The asymptotic properties of these indicators in the setting of semi-supervised learning remains unknown. Moreover, we would also like to investigate the behavior of graph-based semi-supervised learning when the unlabeled data grow faster than the label data. The numerical results suggest that the hard criterion may not be consistent when the size of labeled data grows faster than the size of unlabeled data although the hard criterion still performs constantly better. A theoretical comparison between the two criteria is intriguing under this scenario.


\end{document}